\documentclass[letterpaper, 10 pt, journal, twoside]{ieeetran} 
\usepackage{cite}
\usepackage{soul}
\ifCLASSINFOpdf
 \usepackage[pdftex]{graphicx}
\else
 
\fi
\usepackage{xcolor}

\usepackage{amsmath}
\usepackage{amssymb}
\usepackage{algpseudocode}
\usepackage{algorithm}
 \usepackage{diagbox}
 \usepackage{booktabs}
 \usepackage{multirow}
 \usepackage{multicol}
 \usepackage{array}
 \usepackage{csquotes}
\MakeOuterQuote{"}

\usepackage{amsfonts}
\usepackage{breqn}
\usepackage{amsthm}
\usepackage{mathtools}
\DeclareMathOperator{\diag}{diag}

\usepackage[hidelinks,colorlinks=true,linkcolor=blue,citecolor=blue]{hyperref}
\usepackage{etoolbox}
\makeatletter
\patchcmd{\@makecaption}
  {\scshape}
  {}
  {}
  {}
\makeatother

\usepackage{scalerel}
\newtheoremstyle{upright}
{3pt}{3pt}{\normalfont}{}{\bfseries}{.}{0.5em}{}
\theoremstyle{upright}
\newtheorem{theorem}{Theorem}

\begin{document}

\setlength{\abovedisplayskip}{3pt}
\setlength{\belowdisplayskip}{3pt}

\title{Task-Adaptive Admittance Control for Human-Quadrotor Cooperative Load Transportation with Dynamic Cable-Length Regulation}

\author{{\color{black}Shuai Li, Ton T. H. Duong, Damiano~Zanotto}

\thanks{{\color{black}Manuscript received: February 12, 2026; Accepted: April, 13, 2026.}}
\thanks{{\color{black}This paper was recommended for publication by Editor Cosimo Della Santina upon evaluation of the Associate Editor and Reviewers' comments. (Corresponding author: Damiano Zanotto)}}

\thanks{{\color{black}SL, TD, and DZ ({\tt{dzanotto@stevens.edu}}) are with the Dept. of Mechanical Engineering, Stevens Institute of Technology, Hoboken, NJ 07030, USA. }}

\thanks{{\color{black}Digital Object Identifier (DOI): see top of this page.}}
}
\markboth{IEEE Robotics and Automation Letters. Preprint Version. Accepted April, 2026}%
{Li \MakeLowercase{\textit{et al.}}: Task-Adaptive Admittance Control for Human-Quadrotor Cooperative Load Transportation}

\maketitle

\begin{abstract}
The collaboration between humans and robots is critical in many robotic applications, especially in those requiring physical human-robot interaction (pHRI). Previous research in pHRI has largely focused on robotic manipulators, employing impedance or admittance control to maintain operational safety. Conversely, research in human-quadrotor cooperative load transportation (CLT) is still in its infancy. This letter introduces a novel admittance controller designed for safe and effective human-quadrotor CLT using a quadrotor equipped with an actively-controlled winch. The proposed method accounts for the system’s coupled dynamics, allowing the quadrotor and its cable to dynamically adapt to contact forces during CLT tasks, thereby enhancing responsiveness. We experimentally validated the task-adaptive capability of the controller across the entire CLT process, including in-place loading/unloading and load transporting tasks. To this end, we compared the system performances against a conventional approach, using both variable and fixed cable lengths under low- and high-stiffness conditions. Results demonstrate that the proposed method outperforms the conventional approach in terms of system responsiveness and motion smoothness, leading to improved CLT capabilities.
\end{abstract}

\IEEEpeerreviewmaketitle

\section{Introduction}
\IEEEPARstart{R}ecent advances in autonomous unmanned aerial vehicles (UAVs) and evolving regulations have paved the way for the use of drones in applications requiring frequent and close interactions between humans and UAVs, such as last-mile delivery services \cite{ WingExpands, lmd2025}. Multi-rotor platforms with suspended-load mechanisms leverage agility and active cable-length control to facilitate fast delivery while avoiding unnecessary landing and takeoff. However, suspended-load operations introduce safety concerns, including potential load collisions during docking in unstructured environments and unexpected behavior during interactions with users. Recent reports of drone crashes have heightened safety concerns among customers and regulators \cite{amazonDroneCrash,bloombergDorneSafety}, underscoring the need for methods that ensure safe and reliable human–quadrotor load transportation.\\
\begin{figure}[t!]
\centering
    \includegraphics[width=0.95\linewidth]{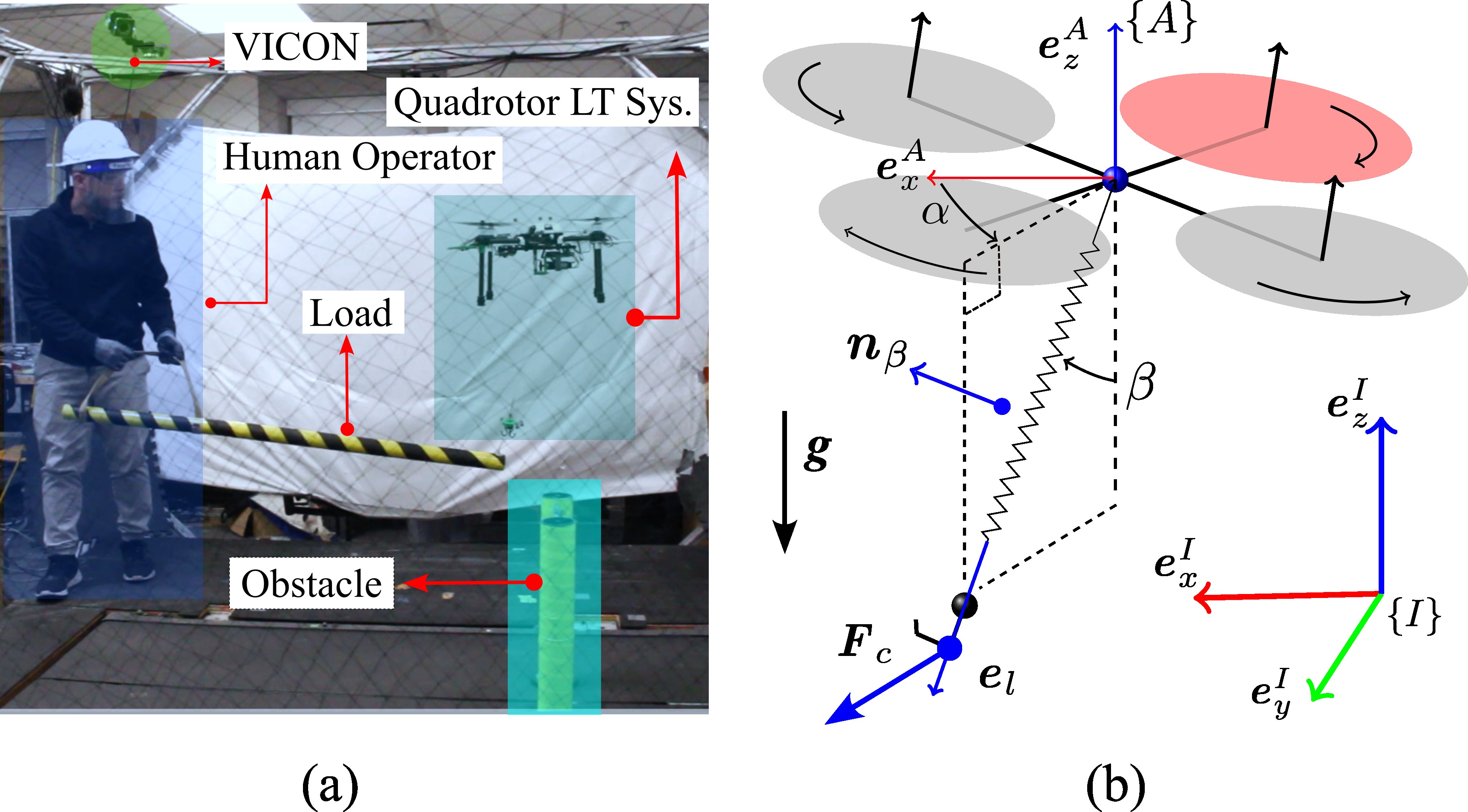}
    \caption{\rmfamily \textbf{(a)} Proposed human-quadrotor CLT system with active cable length control. The obstacle (truncated in height, for better visualization), is positioned along the load's path to the destination, prompting the human operator to take action to alter the system's motion to avoid the obstacle. This is safely achieved using an admittance controller
    . \textbf{(b)} Model of the CLT system, showing the inertial frame $\{I\}$, with coordinate axes along unit vectors $\textbf{\textit{e}}_x^I,\textbf{\textit{e}}_y^I, \textbf{\textit{e}}_z^I$, and the quadrotor-attached 
    frame $\{A\}$. 
    Frame $\{A\}$ is centered in the drone center of mass (COM), with axes parallel to $\{I\}$. 
    $\textbf{\textit{e}}_l$ denotes the unit vector parallel to the cable and pointing away from the drone; 
    $\alpha$ and $\beta$ are the azimuth and inclination angles of $\textbf{\textit{e}}_l$, respectively; $\textbf{\textit{F}}_c$ is the interaction force resulting from the combined effects of the load weight/inertia and the human guiding action; $\textbf{\textit{g}}$ indicates the gravity vector. {\color{black}Video available at:} \url{https://youtu.be/f3Z9GKF7Zgs}}
\label{fig_CLT}
\end{figure}
\indent 
Research on aerial robots capable of physical interaction has advanced rapidly in recent years. Aerial manipulators—combining the agility of UAVs with the dexterity of robotic arms—have demonstrated strong potential in collaborative tasks such as cooperative object transportation \cite{afifi2022toward} and infrastructure inspection \cite{smrcka2021admittance}. These scenarios require the system to regulate interaction forces while maintaining stability during contact. Impedance control establishes a dynamic relationship that maps motion to interaction forces via a virtual mass--spring--damper model \cite{lippiello2018image}, whereas admittance control, its complementary formulation, generates compliant motion in response to external interaction forces and is widely used in aerial systems to support safe physical human--robot interaction \cite{stephens2022aerial,afifi2022toward}.\\
\indent 
Utilizing a cable as a substitute for a manipulator in UAV load transportation represents a promising new 
approach to reduce mechanical complexity and 
facilitate loading and unloading operations during parcel delivery 
\cite{li2019tracking, li2021multi, li2023flight}. While this application requires 
frequent 
physical contact with humans, 
research on how 
to ensure safe interaction in UAV-based parcel delivery systems is scarce. Moreover
, 
the use of an actively-controlled 
cable in human-quadrotor cooperative load transportation (CLT) 
remains unexplored. 
Controllers proposed for aerial manipulators in contact with humans 
cannot be directly extended to the suspended load carrying modality, and only a paucity of studies have explored human-quadrotor collaborative tasks 
involving suspended objects. Tognon \textsl{et al.} \cite{tognon2021physical} proposed a compliant controller that enables a UAV equipped with a cable to guide a human towards a target position, based on the robot's states. In their subsequent work \cite{allenspach2022human}, they incorporated the operator's walking speed in the controller, to synchronize the drone with the operator's pace 
to improve safety. 
Xu \textsl{et al.} \cite{xu2023visual} developed a visual impedance controller for human–UAV collaborative transportation of tethered objects. In their work, the aerial vehicle is controlled to follow the human operator by combining force and vision feedback. Prajapati \textsl{et al.} \cite{prajapati2023aerial} developed a human handle device and a cable attitude sensing device to estimate human commands in 
CLT tasks
, and conducted 
outdoor tests 
for validation. Other groups have carried out initial work investigating the use of admittance control in suspended CLT tasks using multiple UAVs \cite{romano2022cooperative, li2024human}. 
However, the studies referenced in \cite{tognon2021physical,allenspach2022human} do not specifically address the challenges of human-quadrotor CLT tasks, and the methods proposed in \cite{xu2023visual, prajapati2023aerial} may not be applicable to scenarios involving a dynamically changing cable length. In addition, the method proposed in \cite{prajapati2023aerial} offers limited applicability (since it requires estimates of physical attributes of the load, such as size and mass) and limited system autonomy (due to the reliance on explicit human operator commands). Furthermore, the virtual impedance models proposed in previous research 
were {\color{black}based solely on either the motion of the drone or that of the load, without accounting for the coupled dynamics of the CLT system. As a result, these approaches lack the flexibility needed to accommodate a variety of collaborative tasks}.\\
\indent 
In this letter, we introduce a new quadrotor UAV system designed for CLT tasks. The quadrotor features a suspended load-carrying mechanism with an actively controlled cable length. We investigate the coupled dynamics of the system to establish expected contact force-motion relationships in CLT tasks, aiming to enhance the system capacity for efficient response and smooth/compliant motion during physical interactions. The contributions of this work can be summarized as follows: (i) We introduce the first quadrotor system with an actively controlled cable length, designed for human-quadrotor CLT tasks. (ii) We develop a versatile {\color{black}and task-adaptive} model suitable for CLT tasks, accommodating various load shapes and weights. (iii) We propose a novel admittance control strategy, grounded in the 
{\color{black}coupled} 
dynamics of the system, which is effective under both constant and varying cable length (CCL, VCL) conditions. (iv) {\color{black}We present a comprehensive validation of the CLT system through two representative tasks—\textit{loading/unloading in place} and \textit{transporting}—demonstrating the controller’s adaptability and robustness across diverse interaction contexts, and providing a comparative analysis with a conventional approach}.\\
\indent 
The reminder of this letter is organized as follows. Section \ref{sec_model} introduces the dynamic model of the CLT system. Section \ref{sec_acm} describes the 
developed admittance controller and an alternative 
method for comparison. The experimental validation is presented in Section \ref{sec_ExpVal} and results are discussed in Section \ref{sec_diss}. Finally, the paper is concluded in Section \ref{sec_con}.

\section{System Description and Modeling}
\label{sec_model}
We assume a CLT scenario wherein a 
quadrotor-based load transportation system  and a human operator work together to transport a tethered load (Fig.~\ref{fig_CLT}(a)). The quadrotor is equipped with an actuated winch and a load cell mounted underneath its frame. The human operator guides the CLT system by controlling the contact force $\textbf{\textit{F}}_c$ exerted on the hook attached to the distal extremity of the cable. The dynamic model of the human-quadrotor CLT system is developed under the following simplifying assumptions: i) the cable is massless and inextensible; ii) {\color{black}the hook size is negligible relative to the overall CLT system, such that cable–load contact is assumed to occur at the cable end.} Under these assumptions, the CLT system dynamics can be described as follows \cite{zeng2019geometric, li2023flight}
\begin{equation}
\label{dynamic_eqn}
\mbox{\fontsize{7.5}{5}\selectfont $ 
\begin{split}
    \left(M_q + M_{h}\right) \ddot{\textbf{\textit{x}}}_q &= \left( M_q + M_{h} \right) \textbf{\textit{g}} + \left( \textbf{\textit{I}}_3 - \frac{M_{h}}{M_q} {\color{black}\hat{\textbf{\textit{e}}}^2_l} \right) T \textbf{\textit{R}} \textbf{\textit{e}}_z^{B} - M_{h} \ddot{L}\textbf{\textit{e}}_l \\&+ (\textbf{\textit{I}}_3 + 
    \hat{\textbf{\textit{e}}}_l^2 ) \textbf{\textit{F}}_c  + M_{h} L \boldsymbol{\omega}_l \times 
    (\boldsymbol{\omega}_l \times \textbf{\textit{e}}_l) 
    \\
    \textbf{\textit{J}}_q \dot{\boldsymbol{\Omega}} &= - \boldsymbol{\Omega} \times \textbf{\textit{J}}_q \boldsymbol{\Omega} + \boldsymbol{\tau}_q + \boldsymbol{\tau}_m \\
    {\color{black}M_q L^2 \dot{\boldsymbol{{\omega}}}_l }&{\color{black}= -2 M_q L \dot{L} \boldsymbol{\omega}_l - L \textbf{\textit{e}}_l \times T \textbf{\textit{R}} \textbf{\textit{e}}_z^B + \frac{M_q}{M_{h} } L \textbf{\textit{e}}_l \times \textbf{\textit{F}}_c,} 
\end{split}$}
\end{equation}
where $M_q$, $M_{h}$ represent the masses of the quadrotor and hook, {\color{black}respectively}; $\textbf{\textit{x}}_q$, $\boldsymbol{\Omega} \in \mathbb{R}^3$ denote the quadrotor position and angular velocity; $\textbf{\textit{I}}_3 \in \mathbb{R}^{3 \times 3}$ {\color{black}is} the identity matrix; {\color{black} $ \textbf{\textit{g}}$, $\textbf{\textit{e}}_l \in \mathbb{R}^3$ are the gravity vector and the unit vector along the cable, as illustrated in Fig.~\ref{fig_CLT}(b); and $\textbf{\textit{e}}_z^A$, $\textbf{\textit{e}}_z^B \in \mathbb{R}^3$ are the unit vectors along the z-axes of the quadrotor-attached frames $\{A\}$ and $\{B\}$, which are parallel to the z-axis of the inertial frame $\{I\}$ and to the rotation axis of the quadrotor's propellers, respectively (Fig.~\ref{fig_CLT}(b)).} {\color{black}The projection operator $\hat{\cdot} : \mathbb{R}^3 \mapsto  \mathbb{R}^{3\times 3}$} maps a given vector of $\mathbb{R}^3$ to the corresponding skew-symmetric matrix. $T$ represents the quadrotor thrust, while $\textbf{\textit{R}}$ is the rotation matrix describing the orientation of 
$\{B\}$ relative to inertial frame $\{I\}${\color{black}, so that $\dot{\textbf{\textit{R}}} = \textbf{\textit{R}} \hat{\boldsymbol{\Omega}}$}. $L$, $\dot{L}$, and $\ddot{L}$ denote the cable length and its first- and second-order time derivatives. 
$\textbf{\textit{J}}_q \in {\mathbb{R}^{3 \times 3}}$ is the quadrotor's inertia matrix; $\boldsymbol{\tau}_q, \, \boldsymbol{\tau}_m \in {\mathbb{R}^3 }$ represent the torque generated by the quadrotor and the torque exerted by the actuated winch on the quadrotor's body, respectively; $\boldsymbol{\omega}_l \in \mathbb{R}^3$ is the angular velocity of $\textbf{\textit{e}}_l${\color{black}, so that $\dot{\textbf{\textit{e}}}_l = \boldsymbol{\omega}_l \times \textbf{\textit{e}}_l$}. 
We note that, by measuring the contact force $\textbf{\textit{F}}_c$ (which encompasses both the effects of the suspended load and the operator's actions), {we have effectively 
simplified the model and enabled the formulation of a general and versatile equation of motion that does not require knowledge of the physical attributes of the load.}
\vspace{-0.3cm}
\section{Admittance Control}
\label{sec_acm}
\subsection{{\color{black}Coupled} Virtual Impedance Model (CVIM)}
\label{subsec_vim} 
The contact force $\textbf{\textit{F}}_c$ directly affects {\color{black}both} the cable inclination angle $\beta$ and its length $L$, which are regarded as the cable state variables. To make this relationship explicit, we introduce 
vector $\boldsymbol{\eta} = [\alpha, \beta]^T$, where the cable azimuth 
$\alpha$ and inclination 
$\beta$ fully 
determine unit vector 
$\textbf{\textit{e}}_l$ describing the orientation of the cable, as shown in Fig.~\ref{fig_CLT}(b). {\color{black}Starting from the hook dynamics
\begin{equation}
\label{eq:hookdyn}
M_{h} \ddot{\textbf{\textit{x}}}_{h} = \textbf{\textit{F}}_c + M_{h} \, \textbf{\textit{g}} - f_T \, \textbf{\textit{e}}_l,
\end{equation}
we project both sides of \eqref{eq:hookdyn} along the cable direction $\textbf{\textit{e}}_l$. Using this projection, together with  the rotational dynamics of $\textbf{\textit{e}}_l$ given in the last equation of \eqref{dynamic_eqn}, we derive the governing equations for 
$\beta$ and 
$L$ as}
\begin{subequations}\label{vimp_dyn}
\begin{align}
\label{vimpd-rot}
\begin{split}
       M_q M_{h} L^2  \ddot{\beta} &= \left( (-M_{q} M_h L^2 \dot{\textbf{E}} - 2 M_{q} M_h  L \right. \\& \left. \dot{L} \textbf{E} ) \dot{\boldsymbol{\eta}} - M_{h} L \hat{\textbf{\textit{e}}}_l \textbf{\textit{F}}_t + M_q L \hat{\textbf{\textit{e}}}_l \textbf{\textit{F}}_c \right) \cdot \textbf{n}_{\beta}, 
       \end{split}
       \\
\label{vimpd-l}
 M_{h} \ddot{L} &= M_{h} \textbf{\textit{g}} \, {\color{black}\cdot}  \, \textbf{\textit{e}}_l - f_T  + \textbf{\textit{F}}_c  \, {\color{black}\cdot}  \, \textbf{\textit{e}}_l,
\end{align}
\end{subequations}
{\color{black}where $\textbf{\textit{E}} \in \mathbb{R}^{3 \times 2}$ maps the time derivatives of the cable angles to the angular velocity $\boldsymbol{\omega}_l$, according to 
$\boldsymbol{\omega}_l = \textbf{\textit{E}} [\dot{\alpha}, \dot{\beta}]^T$ \cite{li2019tracking}}; $\textbf{\textit{n}}_{\beta}$ denotes the normal vector to the vertical plane containing 
$\textbf{\textit{e}}_l$
, as shown in Fig.~\ref{fig_CLT}(b), {\color{black}$\textbf{\textit{F}}_t = T\textbf{\textit{R}}\textbf{\textit{e}}_z^B$ is the quadrotor thrust vector,} and $f_T$ is the tensile force along the cable, which is governed by the actuated winch
. Equations \eqref{vimp_dyn} 
break down the effects of $\textbf{\textit{F}}_c$ 
into rotational ($\beta$) and translational ($L$) components
, while the other states of the CLT system are determined by 
\eqref{dynamic_eqn}. 
The nominal dynamics are defined as those corresponding to the system operating {\color{black}along a} virtual reference {\color{black}trajectory in the absence of} contact force{\color{black}s. The virtual reference trajectory, denoted by the superscript $v$, represents a user-defined 
trajectory that depends on the task. Thus, the} nominal form of \eqref{vimp_dyn} can be expressed as follows:
\begin{subequations}
\label{nominal_dyn}
\begin{align}
\begin{split}
M_q M_{h} {L^v}^2 \ddot{\beta}^v &= 
        \left ( 
            (-M_q M_{h} {L^v}^2 \dot{\textbf{E}}^v - 2 M_q M_{h} L^v \right. \\& \left.\dot{L}^v \textbf{E}^v) \dot{\boldsymbol{\eta}}^v -M_{h} L^v \hat{\textbf{\textit{e}}}_l^v \textbf{\textit{F}}_t^v
        \right) \cdot \textbf{n}_{\beta}^v, 
        \end{split}
            \\
M_{h} \ddot{L}^v &= M_{h} \textbf{\textit{g}}\, {\color{black}\cdot}  \, \textbf{\textit{e}}_l^v -f_T^v \textbf{\textit{e}}_l^v.
\end{align}
\end{subequations}
{\color{black}After introducing the variations $\delta \beta := \beta - \beta^v$ and $\delta L := L - L^v$, the error dynamics are derived from the coupled dynamics by subtracting \eqref{nominal_dyn} from \eqref{vimp_dyn}, yielding}
\begin{subequations}
\label{error_dynamics}
\begin{align}
M_q M_h {L^v}^2 \delta \ddot{\beta} &= \left( \cdot \right ) + M_q L\hat{\textbf{\textit{e}}}_l \textbf{\textit{F}}_c \, {\color{black}\cdot}  \,\textbf{\textit{n}}_{\beta}\\
M_h \delta \ddot{L} &= \left( \cdot \right) + \textbf{\textit{F}}_c \, {\color{black}\cdot}  \, \textbf{\textit{e}}_l,
\end{align}
\end{subequations}
where $( \cdot)$ denotes additional nonlinear terms. In line with prior research on human–quadrotor CLT that relied on virtual impedance models to regulate the quadrotor response to interaction forces via prescribed inertia, damping, and stiffness coefficients \cite{xu2023visual, li2024human}, this study adopts an impedance-shaping approach to establish a desired compliant CLT behavior. The proposed coupled virtual impedance model (CVIM) is constructed to preserve the system’s physical inertia in \eqref{error_dynamics}, as well as the physical force-transmission channels associated with $\mathbf{\textit{F}}_c$, while neglecting additional 
nonlinear terms $( \cdot)$. By jointly considering deviations in cable inclination and cable length within a unified framework, desired damping and stiffness matrices are introduced to regulate the evolution of $\delta \beta$ and $\delta L$, such that their response to interaction forces follows a prescribed second-order mechanical behavior, leading to the CVIM formulation as follows
\begin{equation}
\label{cvim}
    \textbf{\textit{M}} \ddot{\boldsymbol{\zeta}} + \textbf{\textit{B}} \dot{\boldsymbol{\zeta}} + \textbf{\textit{K}} \boldsymbol{\zeta} = \boldsymbol{\tau}_c,
\end{equation}
where $\textbf{\textit{M}}=\diag(M_q M_{h} {L^v}^2,\, M_h)$ is the inertia matrix, and $\boldsymbol{\zeta} = [\delta \beta,\, \delta L]^T$ is the virtual impedance state vector. {\color{black}$\textbf{\textit{K}} = \diag(K_{\beta}, K_l)$ denotes the positive definite stiffness matrix with gains $K_{*} \in \mathbb{R}^{+}$, and the damping matrix is chosen as $\textbf{\textit{B}} = 2\sqrt{\textbf{\textit{M}} \textbf{\textit{K}}}$ to achieve critical damping. Because the virtual cable length $L^v$ is a time-varying, user-defined reference trajectory, $\textbf{\textit{M}}$ and $\textbf{\textit{B}}$ are also time-dependent. The forcing term is given by $\boldsymbol{\tau}_c = [M_q L \textbf{n}_{\beta} \, {\color{black}\cdot}  \,(\hat{\textbf{\textit{e}}}_l \textbf{\textit{F}}_c), ~\textbf{\textit{F}}_c \, {\color{black}\cdot}  \,\textbf{\textit{e}}_l]^T$.
} 
Thus, 
\eqref{cvim} governs the compliant response of the CLT system in terms of 
$\delta \beta$ and $\delta L$, given a CLT task 
described by a virtual reference profile, in relation to an applied contact force. This 
impedance is 
rendered 
by controlling both the drone's translational motion and the cable's length so that $\beta$ and $L$ are adjusted in response to $\textbf{\textit{F}}_c$.   
\begin{figure}[t!]
    \centering
    \includegraphics[width=0.8\linewidth]{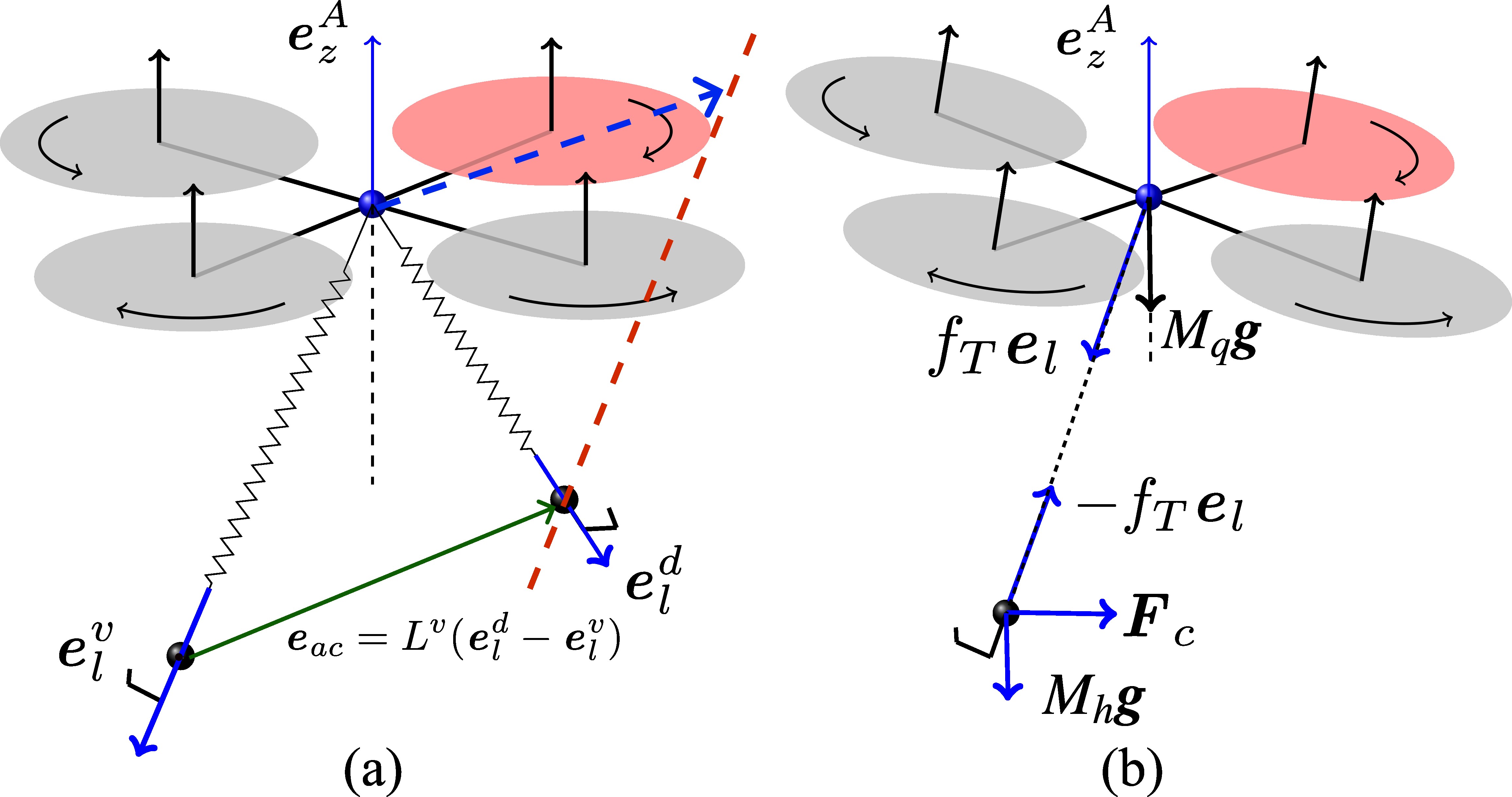}
    \caption{\rmfamily \textbf{(a)} In the {\color{black}coupled} virtual impedance model (CVIM), the admittance error $\textbf{\textit{e}}_{ac}$ is proportional to the difference 
    between the virtual load direction $\textbf{\textit{e}}_l^v$ and its desired direction $\textbf{\textit{e}}_l^d$ {\color{black}(green arrow), which also represents the required quadrotor translational motion (blue dashed arrow). 
    \textbf{(b)} In the simplified virtual impedance model (SVIM), the compliant motion during interaction is separated into two components: the quadrotor translational motion and the cable-length variation, both driven by the tensile force $f_T$.}}
    \label{fig:VIM_S_eac}
    \vspace{-0.5cm}
\end{figure}

\subsection{Compliant Motion Generation}
\label{subsec_CMG}
The desired inclination angle and length for the cable can be derived from \eqref{cvim}, as $\beta^d = \beta^v + \delta \beta$ and $L^d = L^v + \delta L$, respectively. {\color{black}The resulting adjustments to the quadrotor position and cable length are defined as}
\begin{subequations}
    \label{delta_cvim}
    \begin{align}
    \delta \textbf{\textit{x}}_q &= \boldsymbol{\xi}_1 \textbf{\textit{e}}_{ac} + \boldsymbol{\xi}_2 \int_{t_0}^t \gamma^{t - \tau} \textbf{\textit{e}}_{ac} d{\tau} \\
    \delta \bar{L} &= \xi_3 \delta L + \xi_4 \int_{t_0}^{t} \gamma^{t - \tau}\delta L d{\tau},
    \end{align}
\end{subequations}
{\color{black}where $\boldsymbol{\xi}_{1}, \boldsymbol{\xi}_{2} \in \mathbb{R}^{3 \times 3}$ are positive-definite diagonal matrices that determine the drone's motion correction; weights $\xi_{3}, \xi_{4} \in \mathbb{R}^{+}$ are used for compliant cable-length adjustment; $\gamma \in [0,1]$ is the forgetting factor; $t_0$ and $t$ represent the initial and current time, respectively; and $\textbf{\textit{e}}_{ac}$ is the admittance error, defined as} 
\begin{equation}
    \label{eqn_eac}
    \textbf{\textit{e}}_{ac} = L^v (\textbf{\textit{e}}_l^d - \textbf{\textit{e}}_l^v),
\end{equation}
where $\textbf{\textit{e}}_l^d = [\sin{\beta^d} \cos{\alpha^c}, \sin{\beta^d} \sin{\alpha^c}, -\cos{\beta^d}]^T$, $\textbf{\textit{e}}_l^v=[\sin{\beta^v} \cos{\alpha^c}, \sin{\beta^v} \sin{\alpha^c}, -\cos{\beta^v}]^T$, and superscript $c$ indicates the current states. {\color{black} As shown in Fig.~\ref{fig:VIM_S_eac}(a),  $\textbf{\textit{e}}_{ac}$ represents the translational adjustment of the quadrotor required to maintain the cable inclination on the virtual profile while the load remains on the desired trajectory. 
We emphasize that the formulation in \eqref{delta_cvim} represents a command-shaping operation. Unlike the method in \cite{keemink2018admittance}, which introduces a differential feedforward term to improve high-frequency tracking performance, the proposed approach reshapes the virtual admittance by incorporating both a proportional term and a discounted integral term based on recent interaction data, thereby generating physically feasible motions for the quadrotor and cable. The discounted integral term mitigates aggressive motions induced by rapid interaction changes by weighting recent data more heavily, keeping the induced quadrotor motion and cable-length variation close to the recent state. Meanwhile, the proportional term provides a scaled adjustment in response to the current interaction force, maintaining responsiveness to instantaneous inputs. This design is not intended merely to smooth the impedance output; rather, it ensures that the resulting motion commands remain physically trackable by the quadrotor and cable while preserving responsiveness to interaction forces. The }
desired motion of the quadrotor and cable length are computed from the adjustments {\color{black} \eqref{delta_cvim}} and the corresponding reference values $\textbf{\textit{x}}_q^v$ and $L^v$ as follows: 
\begin{equation}
\begin{split}
\label{des_motion}
\textbf{\textit{x}}_q^d &= \textbf{\textit{x}}_q^v + \delta  \textbf{\textit{x}}_q \\
L^d &= L^v + \delta \bar{L}. 
\end{split}
\end{equation}
The overall control architecture of 
CVIM is {\color{black}depicted} in Fig.~\ref{control_diag}. {\color{black}The proposed admittance controller comprises the virtual impedance model and the motion generator (shaded rectangle). Implementation details for the state and force estimation are provided in Sec.~\ref{subsec_force_est}.
\begin{figure}[t!]
\centering 
\includegraphics[width=0.9\linewidth]{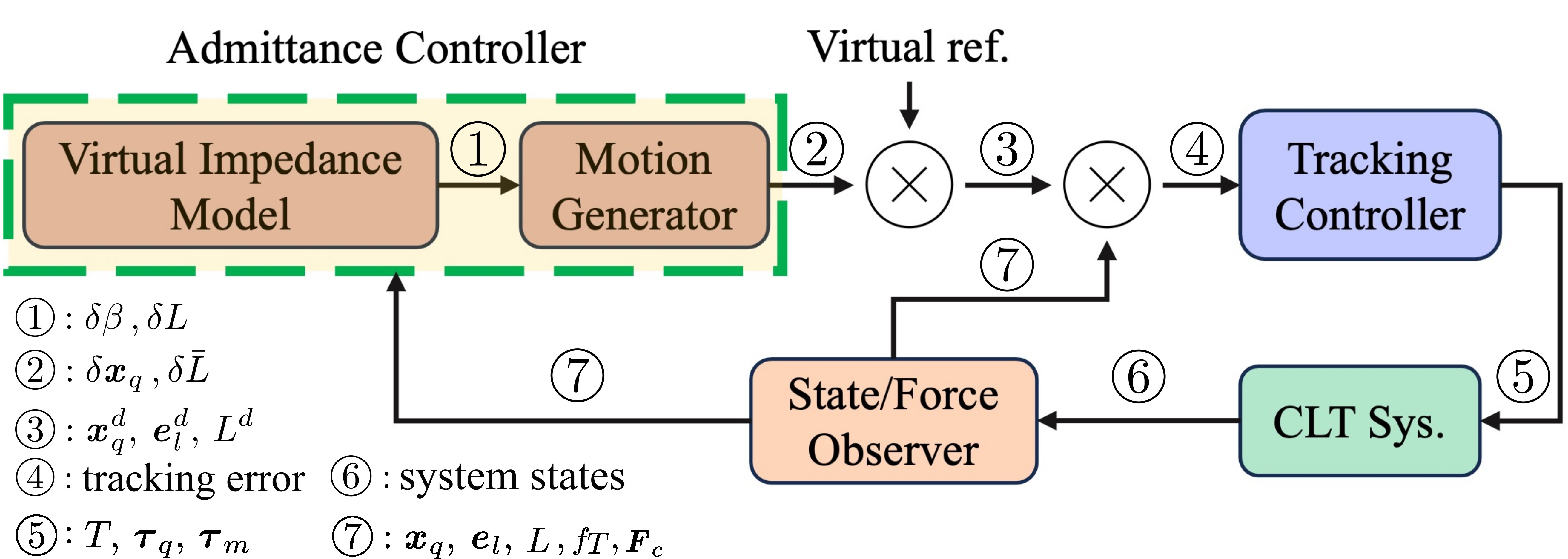}
\caption{\rmfamily Proposed control scheme for the human–quadrotor CLT tasks. {\color{black}Circled numbers indicate the output signals of each block, as described in the legend.}}
\label{control_diag}
\vspace{-0.450cm}
\end{figure}
\subsection{Simplified Virtual Impedance Model (SVIM)}
\label{subsec_vims}
A simplified virtual impedance model (SVIM) was developed as a benchmark to quantify the advantages of the proposed CVIM. The SVIM considers only the translational dynamics of the quadrotor and the cable-length variation under $f_T$
, as illustrated in Fig.~\ref{fig:VIM_S_eac}(b). Following the procedure outlined in Sec.~\ref{subsec_vim} and Sec.~\ref{subsec_CMG}
, the real and nominal translational dynamics of the quadrotor and the corresponding cable-length evolution can be formulated. The SVIM is designed to regulate the variations $\delta \textbf{\textit{x}}_q'$ and $\delta L'$ from the virtual reference profile to exhibit a compliant behavior, according to 
\begin{equation}
\label{svim}
    \textbf{\textit{M}}' \ddot{\boldsymbol{\zeta}}' + \textbf{\textit{B}}' \dot{\boldsymbol{\zeta}}' + \textbf{\textit{K}}' \boldsymbol{\zeta}' = \boldsymbol{\tau}'_c, 
\end{equation}
{\color{black}where $\textbf{\textit{M}}' = \diag(M_q, M_q, M_q, M_{h})$, $\boldsymbol{\tau}'_c = [{f_T \textbf{\textit{e}}_l }^T, -f_T]^T$ are the inertia matrix and forcing term of the SVIM, } $\boldsymbol{\zeta}' = [{\delta \textbf{\textit{x}}_q'}^T, \delta L']^T$ are the virtual impedance states, consisting of the variations $\delta \textbf{\textit{x}}_q'$ and $\delta L'$ from the virtual reference of the quadrotor position and the cable length, respectively, $\textbf{\textit{B}}' = \diag(B_x', B_y', B_z', B_l')$, $\textbf{\textit{K}}' = \diag(K_x', K_y', K_z', K_l')$ are positive definite damping and stiffness matrices defined by gains 
$B'_{*}$ and $K'_{*} \in \mathbb{R}^{+}$. Similar to CVIM, SVIM renders a critically damped mass–spring–damper response through the selection of appropriate damping gains and generates compliant motion based on control laws similar to 
\eqref{delta_cvim} 
\begin{equation}
    \label{delta_svim}
    \begin{split}
        \delta \bar{\textbf{\textit{x}}}_q' & = \boldsymbol{\xi}'_1 \delta \textbf{\textit{x}}_q' + \boldsymbol{\xi}_2^{'} \int_{t_0}^{t}{\gamma'}^{t-\tau}\delta \textbf{\textit{x}}_q' d{\tau}\\
        \delta \bar{L}' &= \xi'_3 \delta L' + \xi'_4 \int_{t_0}^{t} {\gamma'}^{t-\tau} \delta L' d{\tau},
    \end{split}
\end{equation}where $\boldsymbol{\xi}'_{1},\boldsymbol{\xi}'_{2} \in \mathbb{R}^{3\times 3}$ are positive-definite diagonal matrices accounting for the position adjustment of the quadrotor, $\xi'_{3}, \xi'_{4} \in \mathbb{R}^{+}$ are weights for the cable length correction, and $\gamma' \in [0,1]$ indicates the forgetting factor. {\color{black}Akin to \eqref{delta_cvim}, the command-shaping law in \eqref{delta_svim} enables the quadrotor and the cable-length adjustment mechanism to smoothly and effectively track the generated motion commands, thereby exhibiting compliant behavior under the operator’s guidance.} The desired compliant motion resulting from the SVIM 
is given by:
\begin{equation}
\begin{split}
    {\textbf{\textit{x}}_q^d}' &= \textbf{\textit{x}}_q^v + \delta \bar{\textbf{\textit{x}}}_q' \\
    {L^d}' &= L^v + \delta \bar{L}'.
\end{split}
\end{equation}

\subsection{{Stability Analysis}}
\label{subs_stability}
{\color{black}
The admittance system is input-to-state stable (ISS) with respect to $\boldsymbol{\tau}_c$, ensuring bounded responses during interaction and exponential convergence to zero once the interaction force vanishes. Consequently, the outputs $(\delta \textbf{\textit{x}}_q, \delta \bar{L})$ remain bounded and decay to zero, which, together with the quadrotor tracking controller in~\cite{sreenath2013geometric} and a PID controller for cable-length regulation, guarantees bounded tracking errors and stability of the closed-loop CLT system. A detailed stability analysis is provided 
{\color{black}in Appendix~\ref{sec:stab}.}}
%
%
%
\section{Human-Quadrotor CLT Experiments\label{sec_ExpVal}}
\begin{table}[t!]
    \centering
    \caption{\rmfamily Control parameters used for tests with the CVIM and SVIM.} 
    \includegraphics[width=0.85\linewidth]{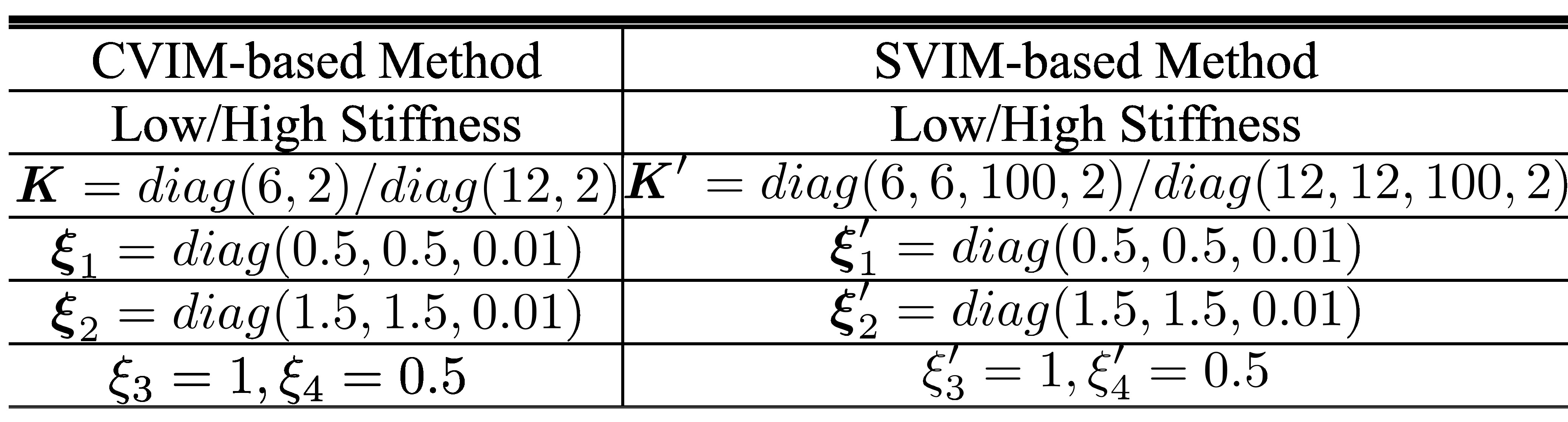}
    \label{tab:ac_param}
    \vspace{-0.3cm}
\end{table}
\subsection{Experimental Setup}
\label{subsec_force_est}
To validate the proposed CVIM, we fabricated a CLT system comprising a quadrotor UAV and a custom-engineered actuated winch (Fig.~\ref{fig_CLT}(a)). {\color{black}Consistent with the modeling assumptions in Sec.~\ref{sec_model}, the masses of the quadrotor and hook were $M_q = 2.1\, \mathrm{kg}$ and $M_h = 0.012\, \mathrm{kg}$, respectively, with corresponding diameters of $0.5\, \mathrm{m}$ and $0.018\, \mathrm{m}$.} The actuated winch featured an optical encoder and a load cell for feedback. {\color{black} A PID controller adjusted t}he cable length 
within the range $\left[0.1; 1.0\right]\, \mathrm{m}$, with $0.5\, \mathrm{m}$ corresponding to a null interaction force. {\color{black}Both t}he admittance controller {\color{black}and the PID were} implemented on a single-board computer (ODROID XU4, Hardkernel co., Ltd., South Korea) running ROS. 
{\color{black}The quadrotor pose $\textbf{\textit{x}}_{q}$ was determined by fusing data from an onboard inertial measurement unit (IMU) and an optical motion capture system (VICON Vero v2.2, Oxford, UK). {The hook position $\textbf{\textit{x}}_{h}$ was obtained {\color{black}from motion capture data sampled at 100Hz.}}} 
The contact force $\textbf{\textit{F}}_c$ was estimated from the hook dynamics  
{\color{black} \eqref{eq:hookdyn}, where the cable orientation $\textbf{\textit{e}}_l$ was determined using motion capture data, and the cable tensile force $f_T$ was estimated from load cell measurements sampled at 10Hz, using a quasi-static force equilibrium model. The hook acceleration $\ddot{\textbf{\textit{x}}}_{h}$ was obtained from the second-order derivative of the hook position interpolated using a cubic spline.}\\ 
\indent
We carried out a set of tests to quantify the advantages of the proposed control method (CVIM) against the baseline method (SVIM). To evaluate the benefits of an actively-controlled cable length, both control methods were tested under two alternative cable configurations (constant cable length (CCL) vs. varying cable length (VCL)). In the CCL configuration, the cable length was maintained {\color{black}at $0.55\, \mathrm{m}$.} For both controllers and both cable configurations, we tested two stiffness levels (low (L) vs. high (H) stiffness). The same human operator carried out ten trials for each combination of \textsl{control method},  \textsl{cable configuration}, and \textsl{stiffness level}, for two representative \textsl{CLT tasks}: Loading/Unloading in Place and Transporting. Across all trials, the cable adjustment was governed by the same stiffness gains, as shown in Tab.~\ref{tab:ac_param}, and the desired altitude of the quadrotor was constrained to a constant value by using a very small gain 
along the $z$ direction. The forgetting factors were set to $\gamma, \gamma' = 0.95$. Other control parameters are reported in Tab.~\ref{tab:ac_param}. All trials 
took place inside a $3.5 \times 3.5 \times 2.8\,\mathrm{m^3}$ indoor cage.
\begin{figure}[t!]
    \centering
    \includegraphics[width=0.85\linewidth]{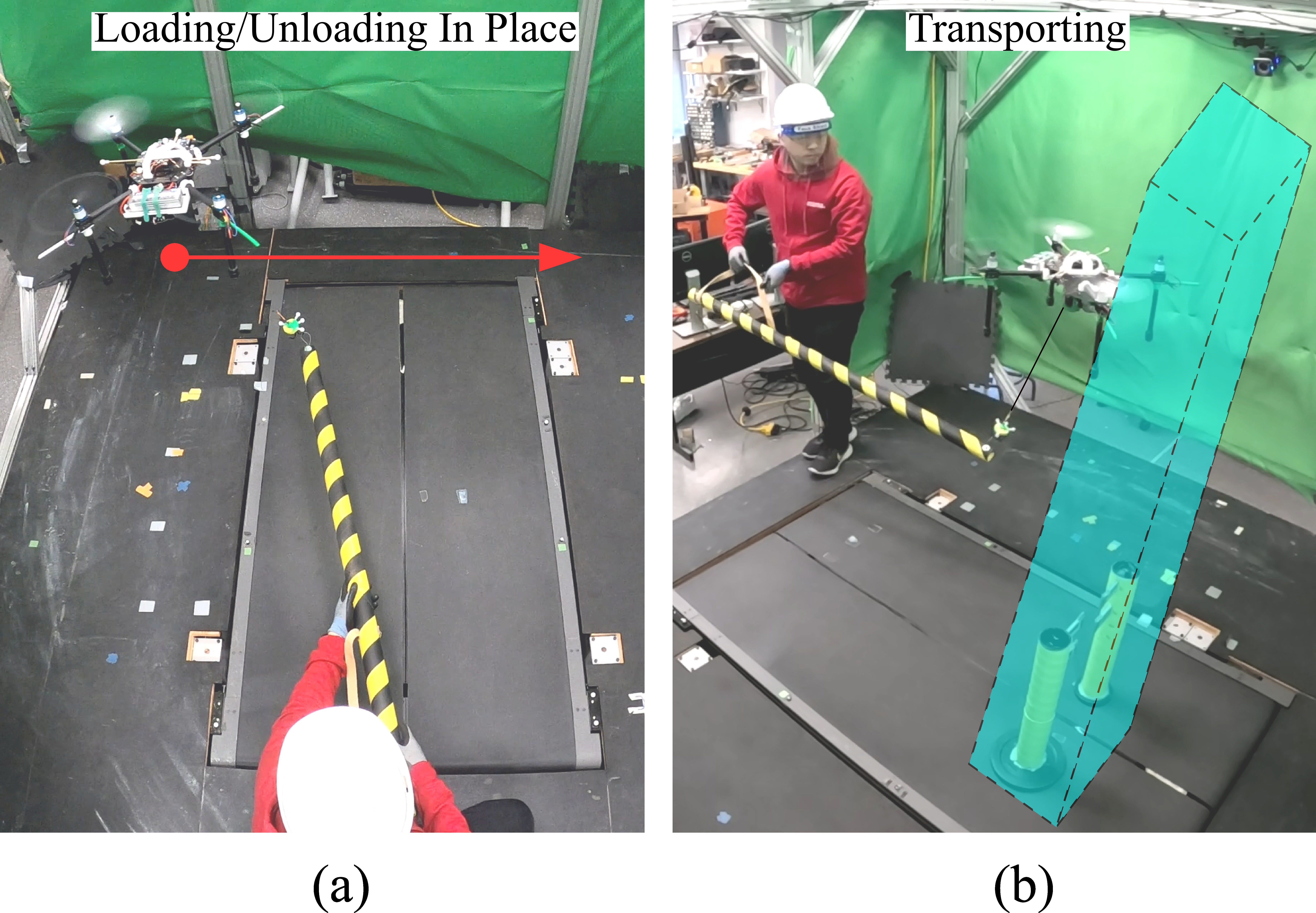}
    \caption{\rmfamily \textbf{(a)} \textsl{Loading/Unloading in Place} – this task requires the operator to attach the load to the cable hook and guide the CLT system to a location as far away as possible from the starting position, within a designated time window; 
\textbf{(b)} \textsl{Transporting} – this task encompasses the collaborative transportation of an object to a target destination, while the operator's action avoids collision between the CLT system and an obstacle located along the desired trajectory, in a position unknown to the quadrotor.}
    \label{fig:two_task}
    \vspace{-0.3cm}
\end{figure}
\\
\indent 
As illustrated in Fig.~\ref{fig:two_task}(a), the Loading/Unloading in Place task simulates a real-world situation where an object needs to be moved to a specific location for unloading, with the assistance of a human operator
.  
The quadrotor 
maintains a hovering state while demonstrating compliance 
to the operator's guiding force. 
In this CLT task, the hovering state serves as a virtual reference point, ensuring that the quadrotor remains stationary and the cable length remains at its initial length when null interaction forces are exerted by the operator. 
The Loading/Unloading in Place task starts with the human operator attaching the load to the cable hook, following which s/he 
is asked to guide the CLT system from its initial position to a location as far away as possible, within a designated time window $\left[t_0; t_f\right]$. Subsequently, the load is detached from the CLT system.\\ 
\indent The Transporting task, shown in Fig.~\ref{fig:two_task}(b), requires the 
CLT system to collaborate with a human operator to transport an object already connected to the cable hook towards a designated destination, following a predefined trajectory (virtual reference). To demonstrate the effectiveness of the compliant controllers in adjusting the motion based on the human guiding force, the operator needs to steer the CLT system to 
navigate around an obstacle (unexpected/unknown to the controller) located along the virtual reference, and return to its designated trajectory.
%
%
%
\subsection{Data Analysis}
Task performance was evaluated in terms of {\color{black}the} traversed distance ({\color{black}path length}) {\color{black}${d}$}, {\color{black} and the integral means of the} 
inclination angle $\bar{\beta}$, 
{\color{black}the} cable tensile force $\bar{f}_T$, and 
{\color{black}the} jerk of the quadrotor trajectory $\bar{J}_q$:
\begin{subequations}
\label{eq:perfmet}
    \begin{align}
        {d} &= \int_{t_0}^{t_f} \parallel \textbf{\textit{v}}_q \parallel dt \\
        \bar{\beta} &= \frac{1}{\left(t_f-t_0\right)} \int_{t_0}^{t_f} \beta \; dt \\
         \bar{f}_T&= \frac{1}{\left(t_f-t_0\right)} \int_{t_0}^{t_f} f_T \; dt \\
         \bar{J}_q &= \frac{1}{\left(t_f-t_0\right)} \int_{t_0}^{t_f} \parallel \ddot{\textbf{\textit{v}}}_q\parallel dt.
    \end{align}
\end{subequations}
In the previous equations, $\textbf{\textit{v}}_q$ denotes the linear velocity of the quadrotor COM. These metrics were selected to provide insights into the CLT system's efficiency and compliance. For the transporting task, the traversed distance ${d}$ 
was normalized to the length of the prescribed trajectory ($2.15 \; \mathrm{m}$), resulting in the dimensionless metric 
$d_n$, which was used in the analysis in place of ${d}$.

{\color{black}For each CLT task, a three-way ANOVA was {\color{black}conducted} to {\color{black}assess the main} effects of \textsl{control method},  \textsl{cable configuration}, and \textsl{stiffness level}, {\color{black}as well as} their interactions, on the performance metrics {\color{black}in \eqref{eq:perfmet}}. {\color{black}Statistical significance was evaluated at} {\color{  black}$\alpha =.05$}. The performance metrics were evaluated using MATLAB (The MathWorks, Natick, MA), while the statistical analysis was carried out in SPSS Statistics v. 29 (IBM, Armonk, NY).}%
\subsection{Results}
\label{sec_exp} 
\subsubsection{Loading/Unloading in Place\label{phase_1}}
\indent 
Figure~\ref{fig:bar_plot_1}(a)-(d) illustrates the mean and standard error (SE) of ${d}$, $\bar{\beta}$, $\bar{f}_T$ and $\bar{J}_q$ for each combination of control method (M), quadrotor configuration (C), and stiffness level (S), evaluated across the trials, for the Loading/Unloading in Place task. Results from the three-way ANOVA are summarized in Table~\ref{tab:anova_task2}(a).\\
\indent
The three-way ANOVA revealed significant M*C and M*S interactions for traversed distance (${d}$), indicating that the effect of the control method on ${d}$ was modulated by both cable configuration and stiffness level. Simple-effect analyses showed that, at both L and H stiffness levels, the CVIM controller significantly increased ${d}$ across both cable configurations ($p<.001$ and $p<.01$ for VCL and CCL, respectively). Actively controlling the cable length had a similar effect of increasing ${d}$ at high stiffness ($p<.01$, $p<.05$ for CVIM and SVIM, respectively), though this effect was observed only for the CVIM controller at low stiffness ($p<.001$).
\begin{figure}[t!]
    \centering
\includegraphics[width=0.85\linewidth]{allGrouped_bar_plot_v5.jpg}
     \caption{\rmfamily \textsl{Loading/Unloading in Place} task: \textbf{(a)} Traversed distance $d$, \textbf{(b)} inclination angle $\bar{\beta}$, \textbf{(c)} smoothness of the drone trajectory $\bar{J}_q$, and \textbf{(d)} cable tensile force $\bar{f}_T$, averaged across ten trials, for each combination of control method, cable configuration, and stiffness level. \textsl{Transporting} task: \textbf{(e)} Normalized travel distance $d_n$, \textbf{(f)} $\bar{\beta}$, \textbf{(g)} $\bar{J}_q$, and \textbf{(h)} $\bar{f}_T$, averaged across ten trials for each combination of the same three factors. Error bars indicate $\pm 1$SE.}
    \label{fig:bar_plot_1}
\end{figure}

\begin{table}[t!]
    \centering
    \caption{\rmfamily {\color{black}Three-way ANOVA tables} for \textbf{(a)} \textsl{Loading/Unloading in Place} task  and \textbf{(b)} \textsl{Transporting} task. M: control method; C: cable configuration; S: stiffness level.
    }
    \includegraphics[width=0.75\linewidth]{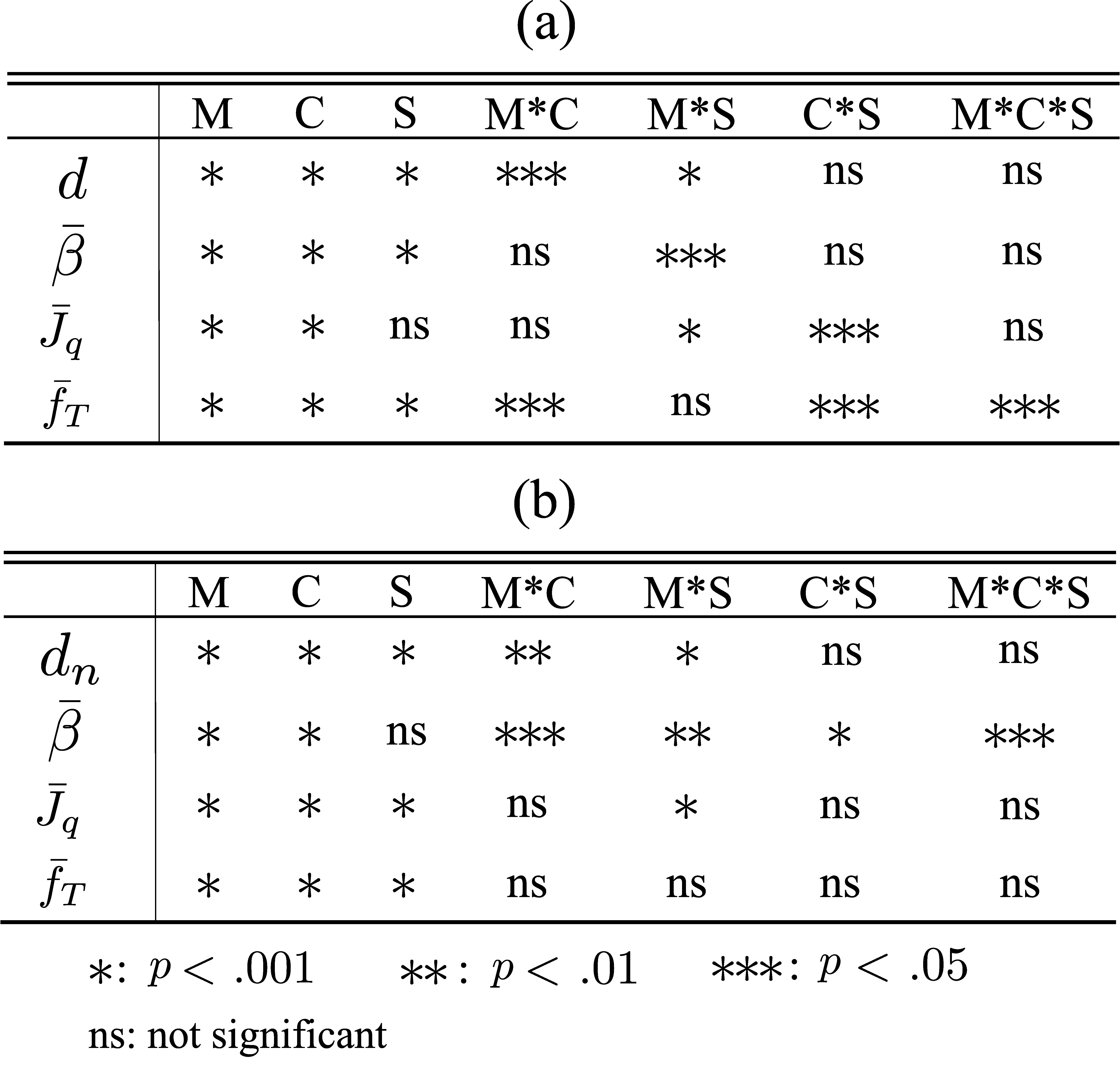}
    \label{tab:anova_task2}
    \vspace{-0.5cm}
\end{table}
Actively controlling the cable length during loading/unloading tasks significantly reduced the inclination angle ($\bar{\beta}$) regardless of control method or stiffness level ($p<.001$). Though the effect of CVIM in reducing $\bar{\beta}$ was more pronounced at low stiffness (significant M*S interaction, $p<.05$), 
follow-up 2-way ANOVAs indicated that the CVIM controller resulted in substantially smaller $\bar{\beta}$, compared to the SVIM controller, at both L and H stiffness levels ($p<.001$ and $p<.05$, respectively). Additionally, lower stiffness resulted in smaller inclination angles ($p<.001$, $p<.01$ for CVIM and SVIM, respectively).

Significant M*S and C*S interactions were observed for trajectory smoothness ($\bar{J}_q$), indicating that the effect of stiffness level on $\bar{J}_q$ during loading/unloading tasks depended on the control method and cable configuration. Follow-up simple-effect analyses showed that a stiffer system increased $\bar{J}_q$, but only for the CVIM controller with a varying-length cable ($p<.001$). Interestingly, the CVIM controller resulted in smoother trajectories at both L and H stiffness levels ($p<.001$ for VCL, $p<.05$ for CCL). Additionally, a varying-length cable resulted in smoother trajectories at both stiffness levels, for both controllers ($p<.001$).

A 3-way interaction was found for the cable tensile force ($\bar{f}_T$). To elucidate the interplay among M, C, and S in affecting $\bar{f}_T$ during loading/unloading operations, separate 2-way ANOVAs were performed for each stiffness level. At high stiffness, the CVIM controller and a varying-length cable independently reduced $\bar{f}_T$ ($p<.001$). However, at low stiffness, the effect of control method depended on the cable configuration ($p<.01$); here, the CVIM controller significantly reduced $\bar{f}_T$ only in the presence of a varying-length cable ($p<.001$).

\indent 
In summary, the above analysis clearly evidenced both independent and synergistic advantages of the proposed CVIM control method, and the dynamically-controlled cable length (VCL) in terms of compliance to the operator's guiding force (increased ${d}$), operational efficiency (decreased $\bar{\beta}$) and safety (smaller $\bar{J}_q$ and $\bar{f}_t$) in loading and unloading CLT tasks.
\newline
\subsubsection{Transporting\label{phase_2}}
Figure~\ref{fig:xy_plot} shows top views of the paths traversed by the quadrotor and by the load during a representative trial for each combination of control method, cable configuration, and stiffness level in the transporting task. 
Compared to the SVIM control method, the CVIM method appears to produce smoother trajectories, with smaller inclination angles – as evidenced by reduced horizontal offsets between the load and the quadrotor paths – and lower interaction forces with the human operator guiding the CLT system, overall allowing for easier operation. {\color{black}S}imilar effects are observed when comparing the use of a varying-length cable (VCL) to a constant-length cable (CCL). Additionally, as stiffness increased, the quadrotor tended to approach the obstacle more closely, even when subjected to greater pulling forces from the operator. 
{\color{black}Figure~\ref{nm_eac_delta_l} shows the norm of the admittance error $\textbf{\textit{e}}_{ac}$ and the cable-length adjustment $\delta\bar{L}$ from the same representative CVIM trials, while Fig.~\ref{fc_cvim} shows the corresponding contact force $\textbf{\textit{F}}_c$, for both VCL and CCL configurations under L and H stiffness levels. At the beginning of the interaction, the H stiffness yields a larger admittance error, as a stiffer impedance requires the operator to apply greater pulling forces to induce motion, resulting in a larger initial cable inclination. As the task progresses, 
the admittance error converges to zero in all cases, indicating a return to the vertical reference configuration. The cable-length variations span similar ranges for both stiffness levels, and the cable remains slightly extended at the end of the task due to tensile forces from the hanging load and residual operator input. Across these representative trials, the CLT system closely tracked the prescribed compliant motion: the quadrotor position-tracking error norm was $0.09\pm0.36\,\mathrm{m}$ (L) and $0.10\pm0.02\,\mathrm{m}$ (H) for VCL, and $0.06\pm0.03\,\mathrm{m}$ (L) and $0.05\pm0.02\,\mathrm{m}$ (H) for CCL; for VCL, the cable-length tracking error was $0.34\pm0.26\,\mathrm{cm}$ (L) and $0.33\pm0.24\,\mathrm{cm}$ (H).}\\
\begin{figure}[t!]
    \centering
    \includegraphics[width=0.85\linewidth]{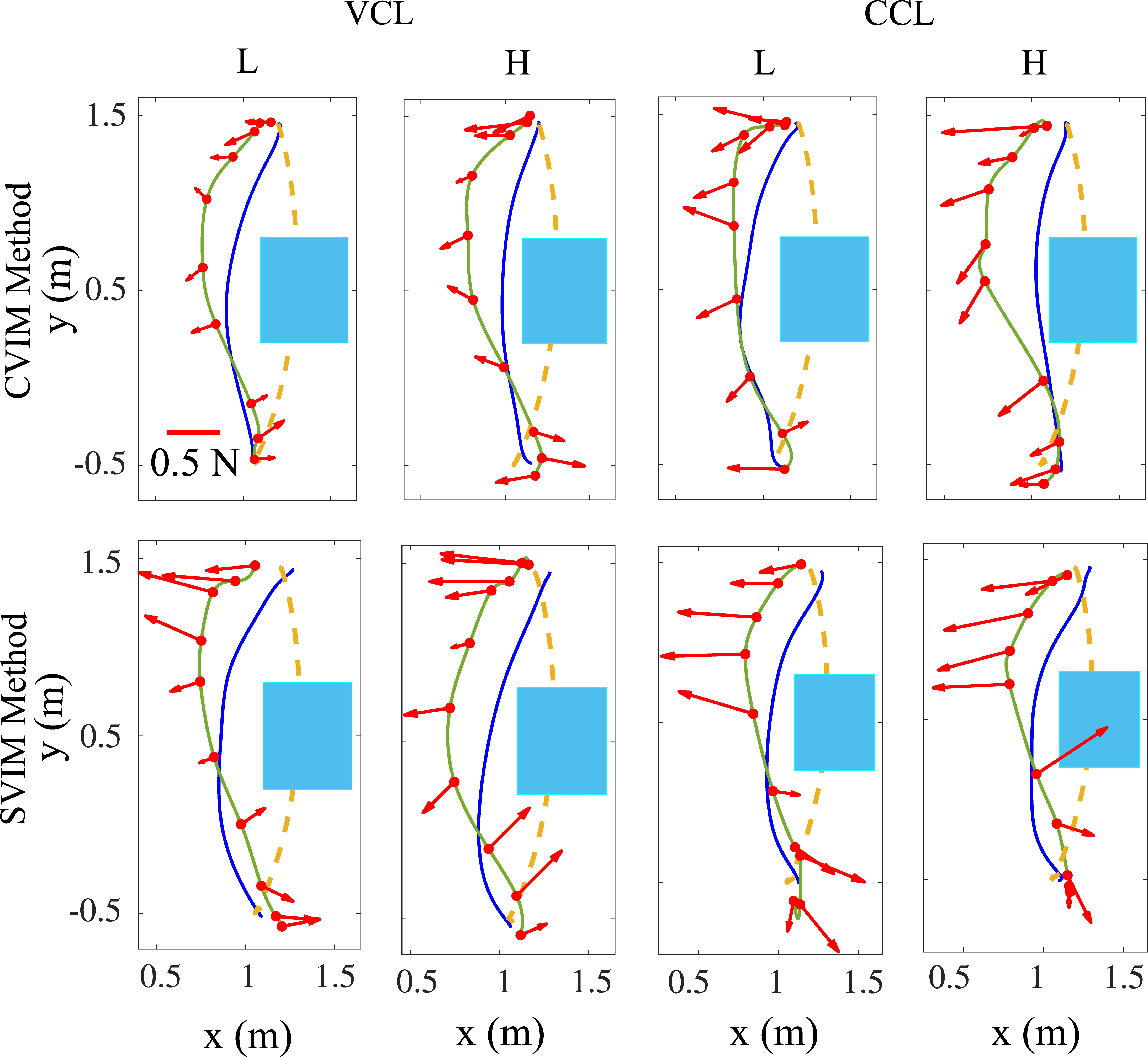}
    \caption{\rmfamily Top views of the {\color{black} virtual references} (dashed orange lines) and operator-guided paths (solid blue lines) followed by the quadrotor's center of mass during representative transport tasks. Green solid lines and red arrows indicate the actual path traversed by the load (represented by the cable's distal point) and the contact force $\textbf{\textit{F}}_c$ exerted by the operator to avoid collisions with the obstacle (light blue rectangles). Representative trials are shown for the CVIM (top row) and SVIM (bottom row) control methods, across all combinations of cable configurations and stiffness levels.
    }
    \label{fig:xy_plot}
\end{figure}
\begin{figure}[t!]
    \centering
    \includegraphics[width=0.80\linewidth]{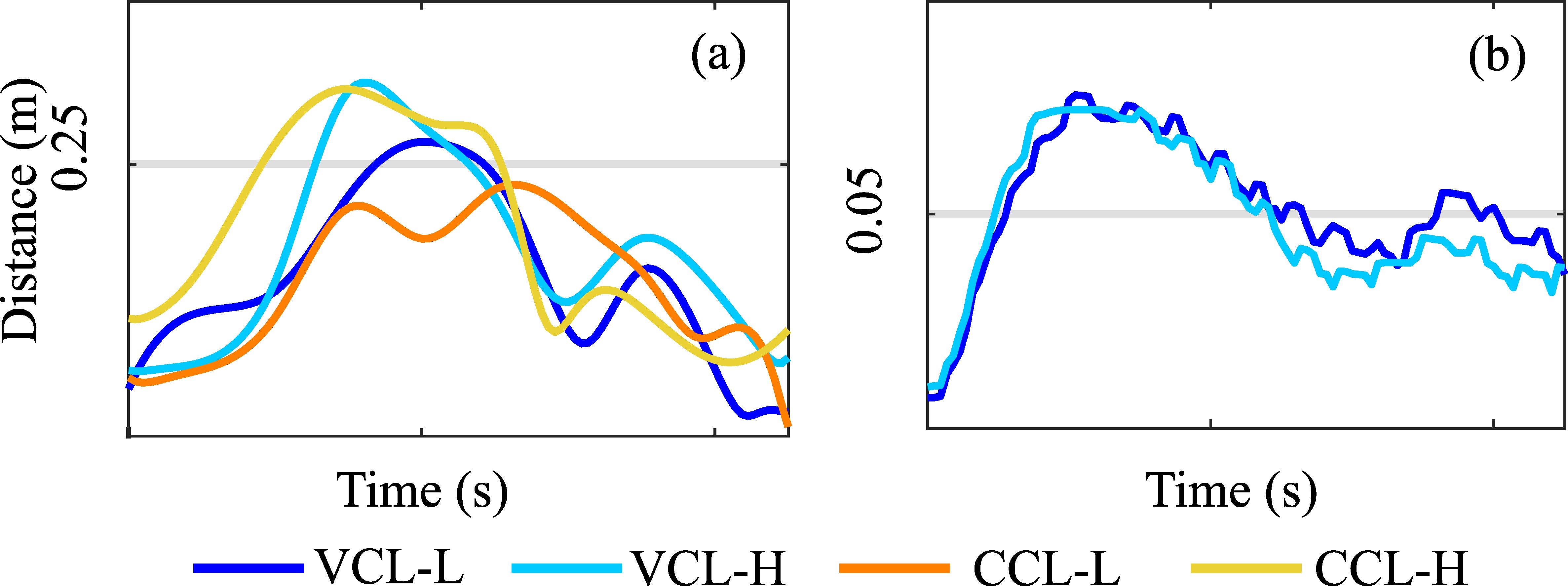}
    \caption{\rmfamily{\color{black} \textbf{(a)} Norm of the admittance error $\textbf{\textit{e}}_{ac}$ for CVIM, under each combination of cable configuration and stiffness level; \textbf{(b)} Corresponding cable-length adjustment $\delta\bar{L}$ for the VCL configurations.}}

    \label{nm_eac_delta_l}
\end{figure}
\begin{figure}[t!]
    \centering
    \includegraphics[width=0.85\linewidth]{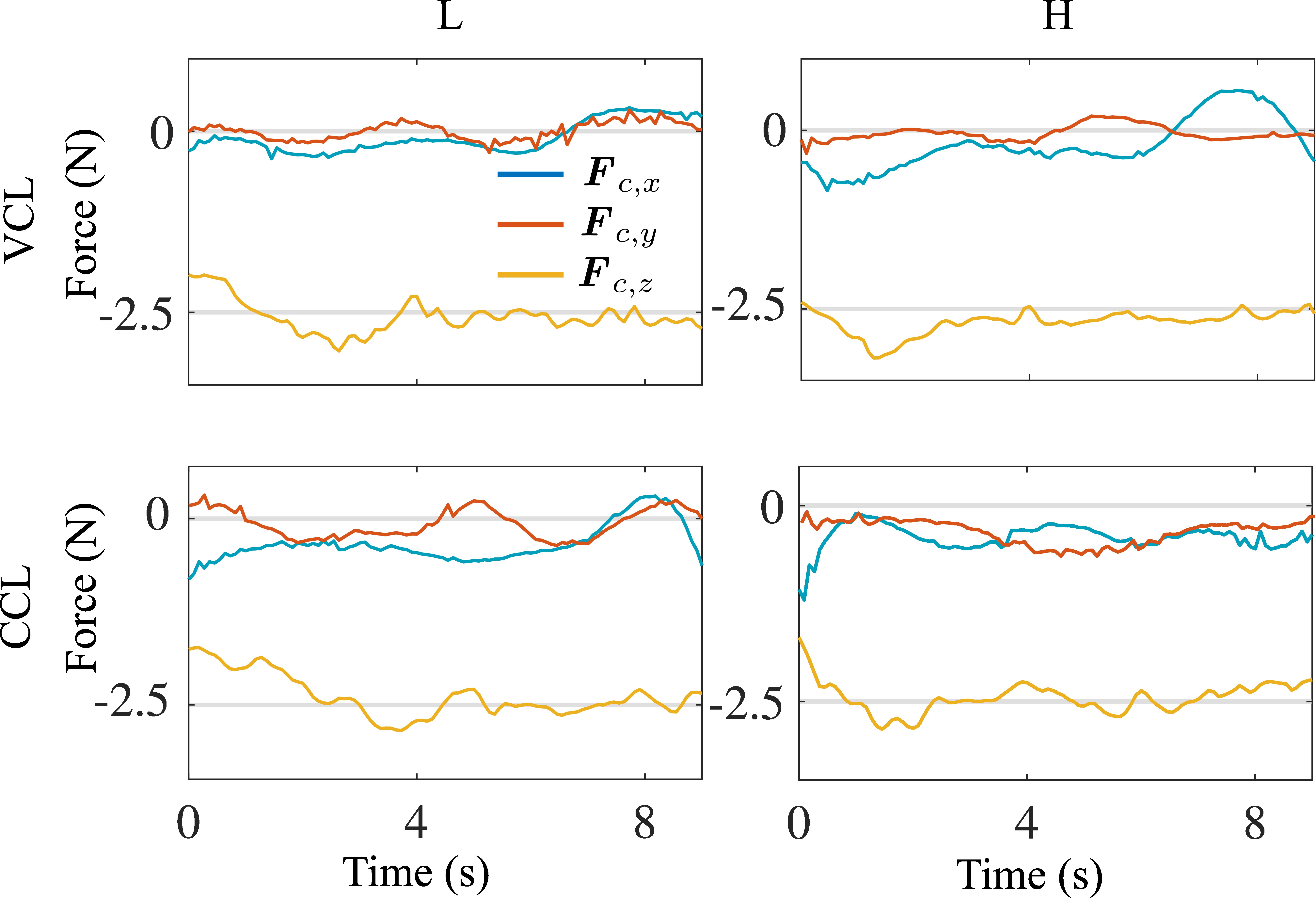}
    \caption{\rmfamily  {\color{black}Contact force $\textbf{\textit{F}}_c$ obtained with CVIM for each cable configuration and stiffness level. The components of $\textbf{\textit{F}}_c$ are reported along the axes of the inertial frame $\{I\}$ defined in Fig.~\ref{fig_CLT}.
    }}
    \label{fc_cvim}
\end{figure}
The three-way ANOVA results for the transporting task are summarized in Table~\ref{tab:anova_task2}(b) and the corresponding performance metrics  are presented in Fig.~\ref{fig:bar_plot_1}(e)–(h). 
For normalized traversed distance ($d_n$), the ANOVA showed significant M*C and M*S interactions. Simple-effect analyses indicated that, under CVIM, the varying-length cable significantly increased $d_n$ at both stiffness levels ($p<.01$), whereas the high stiffness reduced $d_n$ ($p<.05$). No significant cable-length or stiffness effects were observed under SVIM. For the average cable inclination ($\bar{\beta}$), a significant three-way interaction was found. Follow-up two-way ANOVAs showed that, under low stiffness, both CVIM and a varying cable length independently reduced $\bar{\beta}$ ($p<.001$, $p<.05$ for M and C) but under high stiffness, a significant M*C interaction was observed ($p<.01$), with CVIM reducing $\bar{\beta}$ only when combined with a varying-length cable ($p<.01$). Trajectory smoothness ($\bar{J}_q$) improved with varying cable length regardless of control method or stiffness level ($p<.001$). The effect of control method on $\bar{J}_q$ depended on stiffness level. Follow-up two-way ANOVAs indicated that CVIM reduced $\bar{J}_q$ relative to SVIM only at high stiffness ($p<.001$). Additionally, low stiffness resulted in smoother trajectories ($p<.001$, $p<.05$ for CVIM and SVIM). The interaction force ($\bar{f}_T$) decreased with CVIM, with a varying-length cable, and with low stiffness ($p<.001$), with each factor acting independently.\\
\indent
In summary, consistent with the loading/unloading tests, CVIM and the varying-length cable (VCL) improved task performance—independently or synergistically—by increasing compliance ($d_n$), reducing load swing ($\bar{\beta}$), and producing smoother, safer motions (lower $\bar{J}_q$ and $\bar{f}_T$) during human–quadrotor CLT transportation.
\section{Discussion}
\label{sec_diss}
This study introduced a novel admittance control framework (CVIM) for human–quadrotor CLT systems. When combined with a dynamically variable cable length (VCL), the CVIM seamlessly modulated cable inclination and length in response to estimated interaction forces during CLT operations. This resulted in lower interaction forces, smoother quadrotor trajectories, and reduced cable swing angles—enhancing operational efficiency and safety, and ultimately contributed to a more user-friendly experience during both loading/unloading and transporting tasks. 
Importantly, these enhancements were enabled by the use of an actuated winch with embedded load cell for in-flight cable length dynamic modulation and cable force estimation, which had not been explored in previous related works 
\cite{xu2023visual, prajapati2023aerial}.

Unlike recent works that oversimplified the force-motion relationship in CLT tasks by 
neglecting the effects of the swinging 
load \cite{xu2023visual, li2024human}, 
the proposed CVIM takes 
advantage of the system's coupled dynamics to render the causal relations between the physical interaction with the operator and the produced motions. This resulted in 
a CLT system that demonstrates better operational efficiency (lower {\color{black}$\bar{\beta}$}) and exhibits {\color{black}safer} (smaller $\bar{J_q}${\color{black}, $\bar{f}_T$}), and more compliant (larger ${p}$, $p_n$) quadrotor motions, suggesting an 
improved ability to 
{\color{black}collaborate} with the human operator. {\color{black} The proposed admittance controller exhibits task-adaptive autonomy that goes beyond tethered guidance by a human operator \cite{tognon2021physical, allenspach2022human}, enabling more flexible and robust handling of diverse human–quadrotor collaborative load transportation tasks.} In contrast to the aforementioned works, we conducted an in-depth experimental analysis comparing the proposed CVIM approach to a simpler model (SVIM). This comparison enabled us to isolate the individual contributions of the impedance model, cable-length modulation, and stiffness levels to CLT task performance, as well as examine their interactions.\\
\indent
This work aligns with recent efforts aimed at developing new admittance control strategies to enhance safety and effectiveness in human–robot collaboration \cite{barakou2024review}. {\color{black}Unlike} 
recent admittance controllers that rely on force estimators \cite{liu2024coordinated,tagliabue2019robust}, {\color{black}our approach directly measures the force transmitted through the cable.} This eliminates the need for complex estimator designs and avoids potential issues associated with estimator divergence. Furthermore, {\color{black}compared to vision-based estimation methods that infer cable length through indirect position computation \cite{sarvaiya2024hpa,xu2023visual}, our approach directly measures the cable length using a motor encoder, offering} greater flexibility in handling both constant and variable cable lengths. 
{\color{black} In summary, the proposed admittance control framework is grounded in direct measurements of interaction force and cable length. It accounts for the 
dynamics of the coupled CLT system and uses a virtual motion profile as a reference. This enables the robot to adapt to human guidance in real time, follow new trajectories when necessary, and otherwise maintain the predefined path.} Together, these features offer a versatile strategy for a wide range of CLT tasks.\\
\indent 
This study is not without limitations. First, {\color{black}experiments were conducted in a compact indoor environment, constraining the range of CLT scenarios. Second, the CVIM implementation relied on externally estimated quadrotor and load poses from a motion-capture system, limiting practical deployment. Third, the approach does not account for undesirable physical contacts (e.g., human–cable or human–quadrotor collisions), which is critical for safe payload handover and placement
. Future work will focus on eliminating dependence on external tracking (e.g., via onboard depth sensing for pose estimation and scene awareness) and on developing more advanced interaction-interpretation and contact-detection strategies \cite{gao2025enhancing}.} Additionally, the feasibility of the approach will be evaluated in more complex outdoor CLT scenarios.
\section{Conclusion}
\label{sec_con}
This work introduced a new admittance-based control strategy for human–quadrotor collaborative load transportation with a tethered payload. The controller, termed CVIM, establishes a virtual impedance model grounded in the coupled 
system dynamics, accounting for the impact of human-applied forces on both translational and rotational cable-hook motion to generate compliant quadrotor motion. Its performance was evaluated in two representative collaborative tasks: loading/unloading and transporting—under varying- and constant-cable length configurations and low/high stiffness levels, and compared against a conventional admittance controller (SVIM). The results demonstrate that CVIM enhances compliance, reduces interaction forces, and produces smoother trajectories, leading to improved safety, efficiency, and task adaptability in human–aerial robot cooperative load-transport scenarios.
{\bibliographystyle{IEEEtran}
\bibliography{quadCLT_ac.bib}
}
\clearpage
\appendices
\section{Stability Analysis\label{sec:stab}}
\subsection{Stability of the Admittance Controller}
\begin{theorem}
\label{theorem_1}
Consider the proposed time-varying admittance dynamics
\begin{equation}
    \textbf{\textit{M}}(t) \ddot{\boldsymbol{\zeta}} + \textbf{\textit{B}} (t)\dot{\boldsymbol{\zeta}} + \textbf{\textit{K}} \boldsymbol{\zeta} = \boldsymbol{\tau}_c,
\end{equation}
where $\textbf{\textit{M}}(t) = \diag(M_qM_h(L^v(t))^2, M_h)$, $\textbf{\textit{B}}(t) = 2\sqrt{\textbf{\textit{M}}(t) \textbf{\textit{K}}}$, $\textbf{\textit{K}} = \diag(K_{\beta}, K_l)\succ 0 $. The cable length reference $L^v(t)$ satisfies
\begin{equation}
    L^v(t) \in [L^v_{min}, L^v_{max}], ~\mid \dot{L}^v(t) \mid \leq V_L,
\end{equation}
where $L^v_* > 0$ are constant bounds on $L^v(t)$ and $V_L >0$ represents the constant upper bound on the variation rate of the controlled cable length. The forcing vector $\boldsymbol{\tau}_c(t)$ is assumed to be bounded and to vanish after a finite time $T_{c}$:
\begin{equation}
    \Vert \boldsymbol{\tau}_c(t) \Vert \leq \bar{\tau}_c < \infty, \text{for~ all}~t,~  \boldsymbol{\tau}_c(t) \equiv \textbf{0}~ for ~ t \geq T_c.
\end{equation}
Let the command-shaping outputs be
\begin{align}
    \label{cmd_shaping}
    \begin{split}
    \delta \textbf{\textit{x}}_q &= \boldsymbol{\xi}_1 \textbf{\textit{e}}_{ac} + \boldsymbol{\xi}_2 \int_{t_0}^t \gamma^{t - \tau} \textbf{\textit{e}}_{ac} d{\tau}, \\
    \delta \bar{L} &= \xi_3 \delta L + \xi_4 \int_{t_0}^{t} \gamma^{t - \tau}\delta L d{\tau},
    \end{split}
\end{align}
where all gains are positive definite and $0< \gamma < 1$. All symbols and parameters 
retain the definitions given in the main manuscript. Then, the following statements hold:
\begin{enumerate}
    \item The admittance system is input-to-state stable (ISS) with respect to $\boldsymbol{\tau}_c(t)$. Let $\textbf{z} = [\boldsymbol{\zeta}^T, \dot{\boldsymbol{\zeta}}^T]^T$. There exist positive constants $c_1$, $c_2$, $\lambda_1$, and $\Gamma$ such that
    \begin{equation}
    \label{iss}
        \Vert \textbf{z}(t) \Vert^2 \leq \frac{c_2}{c_1}e^{-\lambda t} \Vert \textbf{z}(0) \Vert^2 + \frac{\Gamma}{\lambda c_1}\sup_{0\leq s \leq t}{\Vert \boldsymbol{\tau}_c(s) \Vert^2}.
    \end{equation}

    \item The admittance system is exponentially stable during interaction. If $\boldsymbol{\tau}_c(t) \equiv \textbf{0}~ \text{for} ~ t \geq T_c$, there exists a positive constant $\lambda_2$, such that
    \begin{equation}
    \label{v_exp}
        \Vert z(t) \Vert^2 \leq \frac{c_2}{c_1}e^{-\lambda_2 (t-T_c)} \Vert z(T_c) \Vert^2, ~ t \geq T_c.
    \end{equation}

    \item The command-shaping outputs are bounded for all $t$ and exponentially converge to zero once the interaction vanishes.
\end{enumerate}
\end{theorem}

\begin{proof}
A Lyapunov candidate function is chosen as
\begin{equation}
V_{ac} = \frac{1}{2}\boldsymbol{\zeta}^T\textbf{\textit{K}}\boldsymbol{\zeta} + \frac{1}{2}\dot{\boldsymbol{\zeta}}^T \textbf{\textit{M}}(t)\boldsymbol{\zeta} + \epsilon \boldsymbol{\zeta}^T \textbf{\textit{K}}\dot{\boldsymbol{\zeta}},
\end{equation}
with $\epsilon >0$. Since $\textbf{\textit{M}}(t)$ and $\textbf{\textit{K}}$ are diagonal, $V_{ac}$ can be expressed in block-diagonal form as
\begin{equation}
    V_{ac} = \frac{1}{2}\left( [\delta {\beta}, \delta \dot{\beta}]\textbf{Q}_{\beta}[\delta {\beta}, \delta \dot{\beta}]^T + [\delta L, \delta \dot{L}]\textbf{Q}_L[\delta L, \delta \dot{L}]^T \right),
\end{equation}
where 
\begin{equation}
\textbf{Q}_{\beta} = \begin{bmatrix} K_{\beta}& \epsilon K_{\beta} \\ \epsilon K_{\beta} & M_q M_h {L^v}^2(t)\end{bmatrix}, ~ \textbf{Q}_L = \begin{bmatrix} K_L & \epsilon K_L \\ \epsilon K_L & {\color{black}{M_h}}\end{bmatrix}.
\end{equation}
To ensure $V_{ac} > 0$ uniformly, both $\textbf{Q}_{\beta}$ and $\textbf{Q}_L$ should be positive definite for all $t$. This holds \textsl{iff}
\begin{align}
\begin{split}
    K_{\beta} >0, K_L >0&, ~ \text{and} \\ M_q M_h (L^v(t))^2 - \epsilon^2 K_{\beta} >0&, M_h - \epsilon^2 K_L >0.
\end{split}
\end{align}
Accordingly, let us select an $\epsilon$ satisfying
\begin{equation}
    0 < \epsilon < \min \lbrace \sqrt{\frac{M_qM_h(L^v_{min})^2}{K_{\beta}}}, \sqrt{\frac{M_h}{K_L}} \rbrace,
\end{equation}
so that there exist constants $0<c_1 \leq c_2$ that ensure
\begin{equation}
\label{V_ac_bounds}
    c_1 (\Vert \boldsymbol{\zeta} \Vert^2 + \Vert \dot{\boldsymbol{\zeta}} \Vert^2]) \leq V_{ac} \leq c_2 (\Vert \boldsymbol{\zeta} \Vert^2 + \Vert \dot{\boldsymbol{\zeta}} \Vert^2]).
\end{equation}
Because $V_{ac}$ is uniformly positive definite (under the above conditions) and continuously differentiable with respect to the states and time, it is a valid Lyapunov function. Differentiating $V_{ac}$ and substituting $\ddot{\boldsymbol{\zeta}} = \textbf{\textit{M}}^{-1}(\boldsymbol{\tau}_c - \textbf{\textit{B}} \dot{\boldsymbol{\zeta}} - \textbf{\textit{K}} \boldsymbol{\zeta})$ yields:
\begin{align}
\label{dot_V}
\begin{split}
    \dot{V}_{ac} = &-\dot{\boldsymbol{\zeta}}^T \textbf{\textit{B}} \dot{\boldsymbol{\zeta}} + \frac{1}{2}\dot{\boldsymbol{\zeta}}^T \dot{\textbf{\textit{M}}}\dot{\boldsymbol{\zeta}} + \epsilon \dot{\boldsymbol{\zeta}}^T \textbf{\textit{K}} \dot{\boldsymbol{\zeta}} - \epsilon \boldsymbol{\zeta}^T \textbf{\textit{K}} \textbf{\textit{M}}^{-1}  \textbf{\textit{B}}  \dot{\boldsymbol{\zeta}} \\
    &- \epsilon \boldsymbol{\zeta}^T \textbf{\textit{K}}\textbf{\textit{M}}^{-1}\textbf{\textit{K}}\boldsymbol{\zeta} + \dot{\boldsymbol{\zeta}} \boldsymbol{\tau}_{c} + \epsilon \dot{\boldsymbol{\zeta}}^T \textbf{\textit{K}} \textbf{\textit{M}}^{-1} \boldsymbol{\tau}_{c}.
    \end{split}
\end{align}
Based on the known parameter bounds, let us define:
\begin{align}
\begin{split}
m_{min} &= \min \lbrace M_q M_h (L^v_{min})^2, M_h\rbrace, \\
m_{max} &=  \max \lbrace M_q M_h (L^v_{max})^2, M_h\rbrace, \\
k_{min} &= \min \lbrace \beta, K_L\rbrace, k_{max} = \max\lbrace K_{\beta}, K_L\rbrace,\\
\dot{m}_{max} &= \max_{t}\Vert \dot{\textbf{\textit{M}}}(t) \Vert \leq 2M_qM_h L^v_{max} V_L.
\end{split}
\end{align}
Then,
\begin{align}
\begin{split}
&\lambda_{min}(\textbf{\textit{B}}(t)) \geq b_{min} = 2\sqrt{m_{min} k_{min}}, \\&\lambda_{max}(\textbf{\textit{B}}(t)) \leq b_{max}  = 2\sqrt{m_{max}k_{max}}, \\
& \Vert \textbf{\textit{K}} \textbf{\textit{M}}^{-1}\Vert \leq \frac{k_{max}}{m_{min}}, ~\Vert  \textbf{\textit{K}} \textbf{\textit{M}}^{-1} \textbf{\textit{B}} \Vert \leq 2\frac{k_{max}^{3/2}}{\sqrt{m_{min}}}.
\end{split}
\end{align}
Applying Young’s inequality to bound the cross terms yields positive constants $\eta_{1,2,3} >0$ that satisfy the following relations:
\begin{equation}
\mbox{\fontsize{9}{8}\selectfont $ 
\begin{split}
\vert \epsilon \boldsymbol{\zeta}^T \textbf{\textit{K}} \textbf{\textit{M}}^{-1} \textbf{\textit{B}} \dot{\boldsymbol{\zeta}} \vert &\leq 2\epsilon \eta_1 \frac{k^3_{max}}{m_{min}}\Vert \boldsymbol{\zeta} \Vert + \frac{\epsilon}{2\eta_1} \Vert \dot{\boldsymbol{\zeta}} \Vert, \\
\vert \dot{\boldsymbol{\zeta}}^T \boldsymbol{\tau}_c \vert & \leq \frac{\eta_2}{2}\Vert \dot{\boldsymbol{\zeta}}\Vert^2  + \frac{1}{2\eta_2} \Vert \boldsymbol{\tau}_c\Vert^2, \\
\Vert \epsilon \boldsymbol{\zeta}^T \textbf{\textit{K}} \textbf{\textit{M}}^{-1} \boldsymbol{\tau}_{ac}\vert &\leq \frac{\eta_3}{2} \Vert \boldsymbol{\zeta}\Vert^2 + \frac{\epsilon^2 k^2_{max}}{2\eta_3 m^2_{min}}\Vert \boldsymbol{\tau}_{ac} \Vert^2.
\end{split}  $}
\end{equation}
Next, substituting the above bounds into 
\eqref{dot_V} and collecting terms with respect to $\Vert \boldsymbol{\zeta} \Vert^2$, $\Vert \dot{\boldsymbol{\zeta} }\Vert^2$ and $\Vert \boldsymbol{\tau}_{c} \Vert^2$ yields:
\begin{equation}
\label{d_v_items}
\mbox{\fontsize{7.5}{7}\selectfont $ 
\begin{split}
\dot{V}_{ac} \leq  &-\underbrace{\left(2\sqrt{m_{min} k_{min}} - \frac{1}{2}\dot{m}_{max} - \epsilon k_{max} - \frac{\epsilon}{2\eta_1} - \frac{\eta_2}{2} \right) }_{\mu_2}\Vert \dot{\boldsymbol{\zeta} }\Vert^2 \\
&- \underbrace{\left(\epsilon \frac{k^2_{min}}{m_{max}} - 2\epsilon \eta_1\frac{k^3_{max}}{m_{min}} - \frac{\eta_3}{2} \right)}_{\mu_1} \Vert \boldsymbol{\zeta} \Vert^2 \\
& + \underbrace{\left(\frac{1}{2\eta_2} + \frac{\epsilon^2 k^2_{max}}{2\eta_3m^2_{min}} \right)}_{\Gamma}\Vert \boldsymbol{\tau}_{c} \Vert^2.
\end{split}$}
\end{equation}
To ensure that both coefficients $\mu_1$ and $\mu_2$ are positive, let us select small positive constants $\epsilon$ and $\eta_{1,2,3}$ such that
\begin{align}
\begin{split}
2\sqrt{m_{min}k_{min}} &> \frac{1}{2} \dot{m}_{max} + \epsilon k_{max} + \frac{\epsilon}{2\eta_1} +  \frac{\eta_2}{2}, \\
\epsilon \frac{k^2_{min}}{m_{max}} & > 2\epsilon \eta_1 \frac{k^3_{max}}{m_{min}} + \frac{\eta_3}{2}.
\end{split}
\end{align}
Under these conditions, the following inequality holds:
\begin{equation}
\label{dot_v_simp}
\dot{V}_{ac} \leq -\mu_1 \Vert\boldsymbol{\zeta} \Vert^2 - \mu_2 \Vert \dot{\boldsymbol{\zeta}} \Vert^2 + \Gamma \Vert  \boldsymbol{\tau}_{c}\Vert^2.
\end{equation}
From \eqref{V_ac_bounds} and \eqref{dot_v_simp}, it follows that
\begin{align}
\label{dot_v_min}
\dot{V}_{ac} &\leq -\min(\mu_1, \mu_2)\left( \Vert \boldsymbol{\zeta} \Vert^2+ \Vert \dot{\boldsymbol{\zeta}}\Vert^2 \right) + \Gamma \Vert \boldsymbol{\tau}_{c}\Vert.
\end{align}
Using the quadratic bounds of $V_{ac}$ in \eqref{V_ac_bounds}, we further obtain
\begin{equation}
\dot{V}_{ac} \leq -\frac{\min(\mu_1, \mu_2)}{c_2} V_{ac} + \Gamma \Vert \boldsymbol{\tau}_{c}\Vert = -\lambda_1 V_{ac} + \Gamma \Vert \boldsymbol{\tau}_{c}\Vert,
\end{equation}
where $\lambda_1 = \frac{\min(\mu_1, \mu_2)}{c_2}$. Solving the above differential inequality yields:
\begin{align}
V_{ac}(t) \leq e^{-\lambda_1 t} V_{ac}(0) + \Gamma_1 \int_0^te^{-\lambda_1(t-\tau)}\Vert \boldsymbol{\tau}_{c}\Vert^2 d\tau.
\end{align}
Considering \eqref{V_ac_bounds} and introducing $\textbf{z} = [\boldsymbol{\zeta}^T, \dot{\boldsymbol{\zeta}}^T]^T$, we obtain
\begin{equation}
\label{dot_v_iss}
    \Vert \textbf{z}(t) \Vert^2 \leq \frac{c_2}{c_1}e^{-\lambda_1 t} \Vert \textbf{z}(0) \Vert^2 + \frac{\Gamma}{\lambda_1 c_1}\sup_{0\leq s \leq t}{\Vert \boldsymbol{\tau}_c(s) \Vert^2},
\end{equation}
which proves the ISS property and completes the first part of the theorem.
\\
\indent
When $\boldsymbol{\tau}_{c} \equiv \textbf{0}$ for $ t \geq T_c$, inequality \eqref{dot_v_min} reduces to
\begin{equation}
\dot{V}_{ac} \leq -\lambda_2 V_{ac}, ~ t \geq T_c,
\end{equation}
where $\lambda_2 = \min(\bar{\mu}_1, \bar{\mu}_2)$ and $\bar{\mu}_{1,2}$ denote the coefficients of \eqref{d_v_items} in the absence of the input-related perturbation terms. Integrating it yields \eqref{v_exp}. This completes the second part of the theorem.\\
\indent
From parts 1 and 2 of the theorem, $\delta \beta(t)$ and $\delta L$ are bounded during interaction. Since $\Vert \textbf{\textit{e}}_{ac} \Vert = \Vert  L^v(t)(\textbf{\textit{e}}^d_l - \textbf{\textit{e}}^v_l)\Vert \leq L^v_{max} \vert \delta \beta(t) \vert$, from \eqref{cmd_shaping} we obtain:
\begin{align}
\begin{split}
\Vert \delta \textbf{\textit{x}}_q(t)\Vert &= \Vert \boldsymbol{\xi}_1 \textbf{\textit{e}}_{ac}(t) + \boldsymbol{\xi}_2 \int_{t_0}^t  \gamma^{t - \tau}\textbf{\textit{e}}_{ac}(\tau) d\tau \Vert\\
& \leq \Vert\boldsymbol{\xi}_1\Vert \Vert \textbf{\textit{e}}_{ac}(t) \Vert + \Vert\boldsymbol{\xi}_2\Vert \int_{t_0}^t  \gamma^{t - \tau}\Vert\textbf{\textit{e}}_{ac}(\tau)\Vert d\tau \\
& \leq \Vert\boldsymbol{\xi}_1\Vert \Vert \textbf{\textit{e}}_{ac}(t) \Vert + \Vert\boldsymbol{\xi}_2\Vert \left( \int_{t_0}^t  \gamma^{t - \tau}d\tau\right)\sup_{s\leq t}\Vert \textbf{\textit{e}}_{ac}(s) \Vert\\
& \leq \Vert\boldsymbol{\xi}_1\Vert \Vert \textbf{\textit{e}}_{ac}(t) \Vert + \frac{ \Vert\boldsymbol{\xi}_2 \Vert}{|\ln\gamma|}(1- \gamma^{t-t_0})\sup_{s\leq t}\Vert \textbf{\textit{e}}_{ac}(s) \Vert\\
& \leq \Vert\boldsymbol{\xi}_1\Vert \Vert \textbf{\textit{e}}_{ac}(t) \Vert + \frac{ \Vert\boldsymbol{\xi}_2 \Vert}{|\ln\gamma|}\sup_{s\leq t}\Vert \textbf{\textit{e}}_{ac}(s) \Vert.
\end{split}
\end{align}
Following the same reasoning, for the length adjustment we obtain:
\begin{equation}
\vert \delta \bar{L} \vert \leq \xi_3\vert \delta L(t) \vert + \frac{\xi_4}{\vert \ln(\gamma) \vert}\sup_{s \leq t}\vert \delta L(s) \vert.
\end{equation}
Hence, both $\delta \textbf{\textit{x}}_q$ and $\delta\bar{ L}$ are bounded during interaction and converge to zero once the interaction vanishes. This completes the proof.
\end{proof}

\subsection{Stability of the CLT System based on the Admittance Controller Boundedness}
\begin{theorem}
Given that the outputs of the admittance controller $\delta \textbf{\textit{x}}_q$ and $\delta L$ are bounded during interaction and converge to zero as $\boldsymbol{\tau}_c \rightarrow \textbf{0}$ (Theorem 1), the quadrotor tracking controller in {[20]}{\color{black}, which accounts for the suspended-load effect in the system dynamics, and a conventional PID controller for the cable-length subsystem ensure boundedness of the tracking errors} $\textbf{\textit{e}}_q$ and $e_L$ in the closed-loop CLT system.
\end{theorem}
\begin{proof}
The CLT system dynamics are expressed as described in 
\eqref{dynamic_eqn}. 
Given that virtual references $\textbf{\textit{x}}_q^v$ and $L^v$ are bounded, so are the commanded references: 
\begin{equation}
\textbf{\textit{x}}_q^d = \textbf{\textit{x}}_q^v + \delta \textbf{\textit{x}}_q, ~L^d = L^v + \delta L
\end{equation}
For the quadrotor subsystem, the nonlinear controller {[20]} {\color{black}is designed for the coupled quadrotor–load dynamics, where the load-induced effects are handled within the controller and guarantee convergence of the position tracking error} %
 $\textbf{\textit{e}}_q =\textbf{\textit{x}}_q - \textbf{\textit{x}}_q^d $. \\
For the cable-length subsystem, {\color{black}the winch regulates the scalar cable length independently. Defining $e_L = L - L^d$, the cable-length dynamics are approximated by a double-integrator driven by the winch input. By applying a PID controller, the closed-loop error dynamics are given by}
\begin{equation}
\ddot{e}_L + K_D\dot{e}_L + K_P e_L + K_I \int_0^te_L(\tau)d\tau=0.
\end{equation}
The equilibrium $(e_L, \dot{e}_L, \int_0^1e_L(\tau)d\tau) = (0,0,0)$ is exponentially stable if the gains satisfy (Routh–Hurwitz)
\begin{equation}
K_P >0, K_I >0, K_D >0, K_D K_P > K_I.
\end{equation}
Therefore, all tracking errors remain bounded.
\end{proof}
\end{document}